%% file: acl2023.tex
\pdfoutput=1

\documentclass[11pt]{article}

\usepackage{ACL2023}

\usepackage{times}
\usepackage{latexsym}
\usepackage{arydshln}

\usepackage[T1]{fontenc}

\usepackage[utf8]{inputenc}

\usepackage{microtype}

\usepackage{inconsolata}
\usepackage{booktabs}
\usepackage{graphicx}

\usepackage{xspace}
\usepackage{amsfonts}
\usepackage{amsmath}
\usepackage{booktabs}
\usepackage{cite}
\usepackage{multirow}
\usepackage{graphicx}
\usepackage{listings}

\usepackage{xcolor} %
\definecolor{LightGray}{gray}{0.9}

\definecolor{backcolour}{RGB}{245,248,250}
\definecolor{emph}{RGB}{166,88,53}
\definecolor{nightblue}{RGB}{9,49,105}
\definecolor{keywords}{RGB}{207,33,46}
\definecolor{lightpurple}{RGB}{130,81,223}

\usepackage{arydshln} %
\newcommand{\numposts}{14,441\xspace}

\newcommand{\dataset}{\emph{WhatsThatBook}\xspace}

\newcommand{\ourmethod}{\textsc{MoREL}\xspace}

\title{Decomposing Complex Queries for Tip-of-the-tongue Retrieval}

\author{
    Kevin Lin$^{\spadesuit}$ \quad 
    Kyle Lo$^{\heartsuit}$ \quad 
    Joseph E. Gonzalez $^{\spadesuit}$ \quad 
    Dan Klein$^{\spadesuit}$ \vspace{8pt}\\
    $^\spadesuit$University of California Berkeley 
    \quad 
    $^\heartsuit$Allen Institute for AI \vspace{4pt} \\ 
    \texttt{\small\{k-lin,jegonzal,klein\}@berkeley.edu} 
    \quad 
    \texttt{\small kylel@allenai.org}
}

\begin{document}
\maketitle
\begin{abstract}

When re-finding items, users who forget or are uncertain about identifying details often rely on creative strategies for expressing their information needs---{\textit{complex}} queries that describe content elements (e.g., book characters or events), information beyond the document text (e.g., descriptions of book covers), or personal context (e.g., when they read a book). 
This retrieval setting, called \emph{tip of the tongue} (TOT), is especially challenging for models heavily reliant on lexical and semantic overlap between query and document text.

In this work, we introduce a simple yet effective framework for handling such complex queries by decomposing the query into individual \emph{clues}, routing those as sub-queries to specialized retrievers, and ensembling the results.
This approach allows us to take advantage of off-the-shelf retrievers (e.g., CLIP for retrieving images of book covers) or incorporate retriever-specific logic (e.g., date constraints). We show that our framework incorportating query decompositions into retrievers can improve gold book recall up to 7\% relative again for Recall@5 on a new collection of \numposts real-world query-book pairs from an online community for resolving TOT inquiries.\footnote{Code and data at \href{https://github.com/kl2806/whatsthatbook}{https://github.com/kl2806/whatsthatbook}}

\end{abstract}

\newcommand{\datasetname}{{\textit{WhatsThatBook}}} 

\maketitle

\begin{figure}[t]
\raggedright
\includegraphics[scale=0.45]{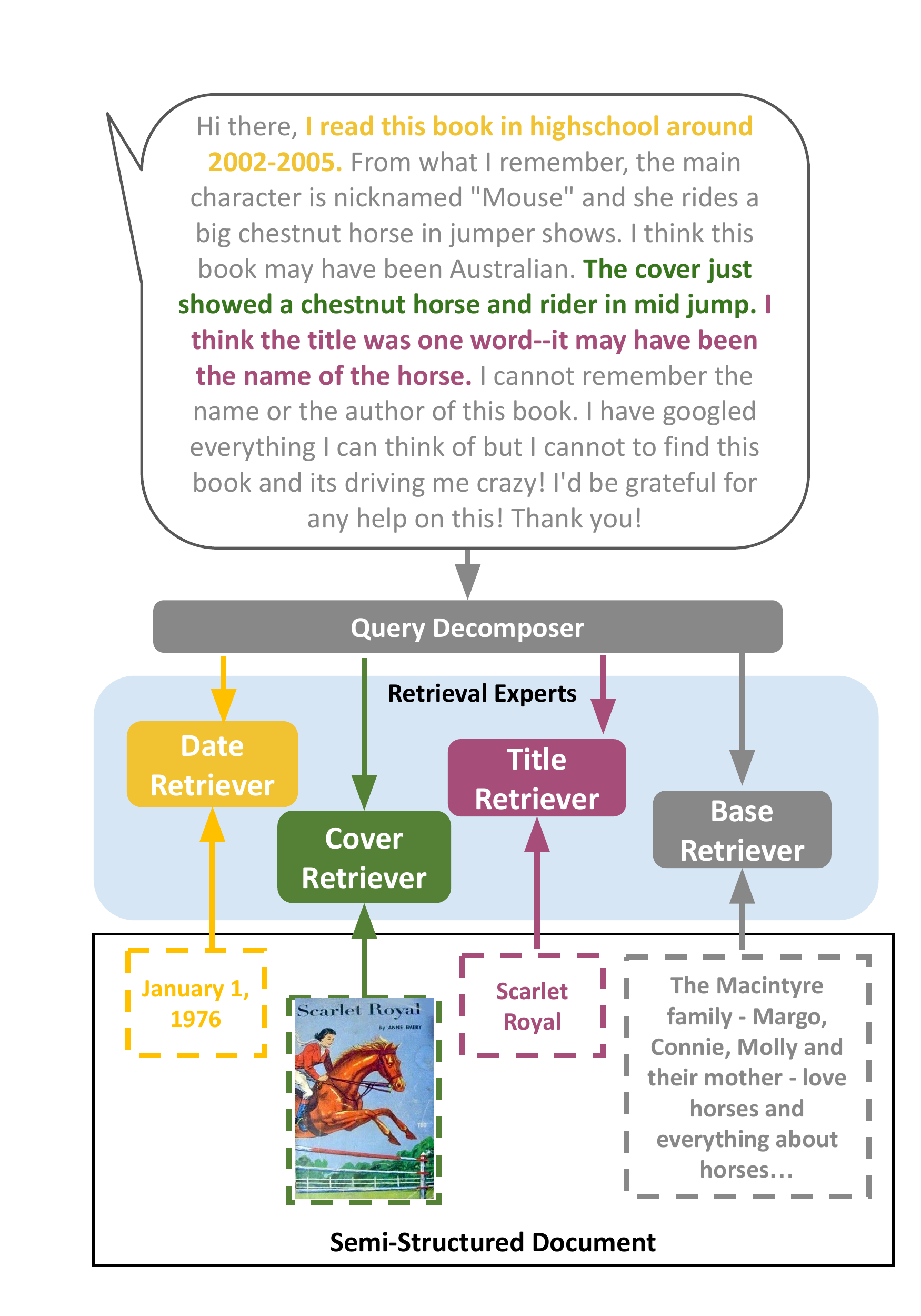}
\caption{\ourmethod decomposes complex queries into subqueries routed to specific retrieval experts.
\vspace{-10pt}}
\label{fig:tipofthetongue}
\end{figure}

\section{Introduction}
Tip of the tongue (TOT) refers to the retrieval setting in which a user is unable to formulate a precise query that identifies a sought item, even if the user knows they've encountered this item before.
For example, users searching for movies they watched or books they read long ago often resort to complex and creative queries that employ a diverse set of strategies to express information relevant to the sought item---high-level categories (e.g., topic, genre), content details from the movie or book (e.g., events, characters), references to personal context (e.g., when they last read the book), descriptions of extratextual elements (e.g., movie promotional posters, book covers), and more. In fact, in an annotation study of TOT queries for movies, \citet{arguello-2021-tot} found over 30 types of informational facets that users may include when crafting queries. Figure~\ref{fig:tipofthetongue} shows a TOT query and its corresponding gold book. 

A key challenge in TOT retrieval is that queries are not just longer and more complex than those in popular retrieval datasets, but resolving them requires an enriched document collection since query-document relevance can't necessarily be established from document content alone (see Table~\ref{tab:dataset_query_length}). For example, in Figure~\ref{fig:tipofthetongue}, the query's description of the book cover---\emph{a chestnut horse and rider in mid jump}---can be highly useful for identifying the book, but necessitates the book's representation to contain that information.

In this work, we present a simple yet effective technique for improving TOT retrieval: First, we augment document representations with additional embeddings derived from additionally linked information (images, metadata). Next, we decompose queries into individual sub-queries or \emph{clues} that each capture a single aspect of the target document. Finally, we route these sub-queries to expert retrievers and combine their results with those from a base retriever that receives the original query.
Experiments show improvement in gold book recall over description-only retrieval baselines on a set of \numposts real-world query-book pairs collected from an online forum for resolving TOT inquiries, complete with cover images and metadata.

\begin{table*}[ht] %
  \centering
  \small
  \begin{tabular}{lll}
    \toprule
    \textbf{Dataset} & \textbf{Query Length} & \textbf{Lexical Overlap}\\
    \midrule
    MSMarco~\citep{Campos2016MSMA-msmarco} & 7.68 & 0.55\\
    Natural Questions~\citep{kwiatkowski2019natural} & 10.35 & 0.52 \\
    BioASQ~\citep{Tsatsaronis2015AnOO-bioasq} & 14.82 & 0.58\\
    TREC-COVID~\citep{trec-covid} & 15.94 & 0.41 \\
    SciFact~\citep{wadden-etal-2022-scifact} & 19.52 & 0.50 \\
    HotPotQA~\citep{yang-etal-2018-hotpotqa} & 22.78 & 0.45 \\
    \midrule
    TOMT~\citep{tomt-bhargav-2022} & 136.50 & 0.25 \\
    \dataset & 156.20 & 0.19\\
    \bottomrule
  \end{tabular}
  \caption{Tip of the tongue (TOT) queries are significantly longer while also having less lexical overlap with the gold document, compared with queries in popular retrieval datasets. Query length is number of BPE~\citep{sennrich-etal-2016-neural-bpe} pieces, averaged across examples. Lexical overlap is fraction of whole words in query that occur in gold passage(s), averaged across examples.}
  \label{tab:dataset_query_length}
\end{table*}

\section{Method}

Given a collection of documents $d_1,\dots,d_n$ and a textual query $q$, the TOT retrieval task aims to identify the sought document $d^*$. The input (raw) documents are semi-structured; each document $d$ contains fields $d^{(1)}, \dots, d^{(k)}$. In the case of books, the fields can correspond to a title, its description, its publication year, an image of its book cover, etc. Missing elements take on a default value (e.g., blank image, earliest publish date in overall book collection). We consider the original document text as one of these fields, which we denote $d^{(o)}$.

\subsection{Query Decomposition}

First, the query decomposer takes a query $q$ and outputs a set of subqueries $q^{(1)}, \cdots, q^{(k)}$. To do this, we use in-context learning with a large language model (LLM) to extract the part of the text from $q$ that is relevant to that field or output the string \texttt{"N/A"} if the $q$ does not contain any relevant information to the field; this is repeated for each field.

In practice, we use GPT~3.5 \texttt{gpt-3.5-turbo} few-shot prompting with in-context 8 examples. An example prompt template (for book covers) is:

\begin{verbatim}
    You are a utility that extracts 
    text related to the cover from 
    a complex query 
    
    Query : { X1 } 
    Cover : { Y1 } 
    
    Query : { X2 } 
    Cover : { Y2 } 
    
    ... 
    
    Query : {X'} 
    Cover :
\end{verbatim}

\noindent where \texttt{X1} through \texttt{X8} are the original query examples for few-shot $q_1, \dots, q_8$, \texttt{Y1} through \texttt{Y8} are gold sub-queries $q_1^{(j)}, \dots, q_8^{(j)}$ (assuming $j$ is the field corresponding to book covers), and the final \texttt{X'} is the  query intended for sub-query extraction.\footnote{\url{https://platform.openai.com/docs/models/gpt-3-5}}
Each field has its own prompt template. Sub-queries for different fields can be generated in parallel, as the they are independent of each other. 

A key implementation detail is that sub-queries need not be pure extractions from the original query. Using LLMs to generate sub-queries affords us the ability to set the few-shot prompt generation targets to be \emph{predictions}. 
This is important as the information in queries are rarely presented in a form amenable for matching with the corresponding document field. For example, books have publish dates, but queries will rarely mention these dates; instead, users may articulate personal context (e.g., ``\emph{I read this book in highschool around 2002-2005}''). Then to simplify the learning task for a date-focused retrieval expert, we might ask the LLM to predict a ``latest possible publish date'' (e.g., 2005).
See Table~\ref{tab:examples} for examples of generated sub-queries.

\subsection{Retrieval Experts} 
\label{sec:retrieval-experts}

We have retriever models, or experts, that specialize to specific field types. Let $R_1, \dots, R_k$ represent these retrievers. Retrievers can be implemented as dense, sparse, or symbolic logic.

If a retriever requires training, we run the query decomposer over all query-document pairs $(q, d)$ in the training set. This produces effectively $k$ training datasets, where each dataset is comprised of a subquery and document-field pair. For example, field $j$ would have training dataset of examples $(q^{(j)}, d^{(j)})$.

At indexing time, each document's field is indexed according to the specifications of its retriever expert. For example, if the retriever is implemented as an embedding model, then that specific field is converted into an embedding. On the other hand, if the retriever is a sparse model, then a sparse index would be built using just that specific field's text.

At inference time, each retriever takes a subquery $q^{(j)}$ and retrieves a document from its associated index of fields.

In practice, for titles and the original book descriptions $x^{(o)}$, we use Contriever~\citep{izacard2021towards}, a state-of-the-art dense retriever.\footnote{\href{https://huggingface.co/facebook/contriever}{https://huggingface.co/facebook/contriever}} For both models, we train for a total of 10,000 steps with a batch size of 16, learning rate of 1e-4. For titles, we finetune with 3,327 extracted sub-queries. For our base retriever, we use the full training set of original book descriptions.

For cover images, we use CLIP~\citep{radford2021learning}, 
a state-of-the-art retriever that can score matches between embedded images and their textual descriptions. Specifically, we finetune ViT-B/32\footnote{\href{https://huggingface.co/sentence-transformers/clip-ViT-B-32}{https://huggingface.co/sentence-transformers/clip-ViT-B-32}} on 2,220 extracted sub-queries using cross-entropy loss with batch size of 4, learning rate of 5e-5 and weight decay of 0.2 for 10 epochs with the Adam optimizer~\citep{kingma2014adam}. We select the model with the best top 1 retrieval accuracy on a validation set.

For publish dates, we use a symbolic function that heuristically scores 0 if a book was published after the sub-query date (i.e. predicted latest publish date) and 1 otherwise. If necessary, we heuristically resolve the sub-query to a year.

\subsection{Combining retrieved results}

In this work, we restrict to a simple strategy of using a weighted sum of all $k$ retrieval scores across the $(q^{(j)}, d^{(j)})$. That is, the final score is:

\[ s(q,d) = \sum_{j=1}^n w^{(j)} R_j(q^{(j)}, d^{(j)}) \]

\noindent All documents are scored in this manner, which induces a document ranking for a given query $q$.

\input{tables/allexamples}

\section{Datasets}
\label{sec:dataset}

We introduce the \dataset dataset consisting of query-book pairs collected from a public online forum on GoodReads for resolving TOT inquiries about books.\footnote{\href{https://www.goodreads.com/group/show/185-what-s-the-name-of-that-book}{https://www.goodreads.com/group/show/185-what-s-the-name-of-that-book}. We scraped data from February 2022.} 
On this forum, users post inquiries describing their sought book and community members reply with links to books on GoodReads as proposed answers.\footnote{This is a simplification of community interactions. Threads also may include dialogue between original poster and members but this is beyond the scope of our work.}
If the searcher accepts a book as the correct answer, the post is manually tagged as \texttt{SOLVED} and a link to the found book is pinned to the thread.
For these solved threads, we take the original inquiry as our query $q$ and the linked book as gold $d^*$. At the end, \dataset contains \numposts query-book pairs. Each query corresponds to a unique book.
Finally, these books are associated with pages on GoodReads, which we used to obtain publication year metadata and images of book covers.

For the experiments in the rest of this paper, we split \dataset{} into train ($n$=11,552), validation ($n$=1,444) and  test ($n$=1,445) sets.  By the nature of our dataset construction, the number of queries and books is equal. We use all 14,441 books, which are gold targets with respect to some query, as our full document collection for indexing.

\section{Experiments}

\input{tables/main-results}

\subsection{Baseline models}
We evaluate our approach against several popular retrieval models that have been used as baselines for a range of other retrieval datsets (see Table~\ref{tab:dataset_query_length}). For text-only models---BM25~\citep{robertson-bm25-1997, robertson2009probabilistic}, Dense Passage Retrieval (DPR)~\citep{karpukhin2020dense}, and Contriever~\citep{izacard2021towards}---the document representation is simply the concatenation of all available document fields into a single text field. For our image-only baseline---CLIP~\citep{radford2021learning}---the document represenntation is only the embedded book cover. All baselines receive the same input (full) query. As well, all baselines are finetuned with the same hyperaparameters as described in \S\ref{sec:retrieval-experts} except using the full training set instead of just examples with a successful sub-query extraction.

\subsection{Results}

Table~\ref{tab:main_results} shows the test results on \dataset{}. We use Recall@$K$ metric as our primary metric since each query has exactly one correct item.

\paragraph{Baselines.} In this setting with low lexical overlap, we see that dense retrievers like DPR and Contriever outperform sparse retrievers like BM25.
Without extracting clues about the book cover, using CLIP on its own is not effective, likely due to its limited context window.\footnote{We pass the full query into CLIP and allow for truncation to happen naturally. This is a big issue with CLIP, which supports a narrow query length; hence, motivating our  approach to extract \emph{clues} about book covers from the full query.} Contriever is the overall best-performing baseline model.

\paragraph{Our method.} Our approach to decompose queries and route clues to specialized retrievers improves performance (ranging from +2 to +3 Recall@$K$ across all cutoffs) over the next best baseline retriever. Looking into the limited expert ablation results, we find that incorporating titles and images that often have more precise descriptions improve the Recall@$K$ for lower values of $K$. Query decomposition improves Recall@5 for images and titles for 3\% to 5\% relative gain respectively. In constrast, the date retriever does not improve recall for lower values of $K$, and instead is more helpful at higher values of $K$. This may be due to the fact that the descriptions are less precise. Incorporating all dates, covers, and titles together provides further gains, indicating the the benefits from each specialized retrieval is somewhat orthogonal and adding additional expert retrievers could be helpful.

\section{Related Work}
\label{sec:related-work}

\paragraph{Dense methods for document retrieval.} 

Document retrieval has a long history of study in fields like machine learning, information retrieval, natural language processing, library and information sciences, and others. Recent years has seen the rise in adoption of dense, neural network-based methods, such as DPR~\citep{karpukhin2020dense} and Contriever~\citep{izacard2022unsupervised-contriever}, which have been shown can outperform sparse methods like BM25~\citep{robertson-bm25-1997, robertson2009probabilistic} in retrieval settings in which query-document relevance cannot solely be determined by lexical overlap. Researchers have studied these models using large datasets of query-document pairs in web, Wikipedia, and scientific literature-scale retrieval settings~\citep{Campos2016MSMA-msmarco, sciavolino-etal-2021-simple, trec-covid}. Many retrieval datasets have adopted particular task formats such as question answering~\citep{kwiatkowski2019natural, yang-etal-2018-hotpotqa, Tsatsaronis2015AnOO-bioasq} or claim verification~\citep{thorne-etal-2018-fever, wadden-etal-2022-scifact}. We direct readers to \citet{Zhao2022DenseTR} for a comprehensive, up-to-date survey of methods, tasks, and datasets.

\paragraph{Known-item and TOT retrieval.}
Tip of the tongue (TOT) is a form of known-item retrieval \citep{buckland-1979, lee-2006-known-item-variations}, a long-studied area in the library and information sciences. Yet, lack of large-scale public datasets has made development of retrieval methods for this task difficult.  Prior work on known-item retrieval focused on constructing synthetic datasets \citep{azzopardi-2007-simulated-queries-european,  kim-croft-2009-pseudo-desktop-simulated, elsweiler-2011-simulated-queries-email-refinding}.
For example, \citet{hagen-2015-known-item-clueweb} released a dataset of 2,755 query-item pairs from \emph{Yahoo!} answers and injected query inaccuracies via hired annotators to simulate the phenomenon of \emph{false memories} \citet{hauff-2011-known-item-query-gen, hauff-2012-realistic-known-item-topics}, a common property of TOT settings.

The emergence of large, online communities for resolving TOT queries has enabled the curation of realistic datasets. \citet{arguello-2021-tot} categorized the types of information referred to in TOT queries from the website \emph{I Remember This Movie}.\footnote{\href{https://irememberthismovie.com/}{https://irememberthismovie.com/}}
Most recently, \citet{tomt-bhargav-2022} collected queries from the \emph{Tip Of The Tongue} community on Reddit\footnote{\href{https://www.reddit.com/r/tipofmytongue/}{https://www.reddit.com/r/tipofmytongue/}} and evaluated BM25 and DPR baselines. 
Our work expands on their work in a key way: We introduce a new method for retrieval inspired by long, complex TOT queries. In order to test our method on a large dataset of TOT queries, we collected a new dataset of resolved TOT queries such that we also had access to metadata and book cover images, which were not part of \citet{tomt-bhargav-2022}'s dataset.

\paragraph{Query Understanding and Decomposition.} Our work on understanding complex information-seeking queries by decomposition is related to a line of work breaking down language tasks into modular subtasks~\citep{andreas2016neural}. More recently, LLMs have been used for decomposing complex tasks such as multi-hop questions into a sequence of simpler subtasks \citep{khot2022decomposed} 
or smaller language steps handled by simpler models \citep{jhamtani2023natural}. 

Related to decomposition of long, complex queries for retrieval is literature on document similarity~\citep{mysore-etal-2022-multi} or query-by-document (QBD)~\citep{Yang2018RetrievalAR}. In these works, a common approach is decomposing documents into sub-passages (e.g. sentences) and performing retrieval on those textual units. The key differentiator between these works and ours is that document similarity or QBD are inherently symmetric retrieval operations, whereas our setting requires designing approaches to handle asymmetry in available information (and thus choice of modeling approach or representation) between queries and documents. In this vein, one can also draw parallels to \citet{lewis-etal-2021-paq}, which demonstrates that retrieving over model-generated question-answering pairs instead of their originating documents can improve retrieval, likely due to improved query-document form alignment. In a way, this is similar to our use of LLMs to generate clues that better align with extratextual document fields, though our work is focused on query-side decomposition rather than document-side enrichment.

\section{Conclusion}
We study a real-world information-seeking setting---tip of the tongue retrieval---in which users issue long, complex queries for re-finding items despite being unable to articulate identifying details about those items. We introduce a simple but effective approach to handling these complex queries that decomposes them into sub-queries or clues that are routed to expert retrievers for specialized scoring. Our simple framework allows for modular composition of different retrievers and leveraging of pretrained models for specific modalities such as CLIP for document images. We observe improvements of up to 7\% relative gain for Recall@$5$ when incorporating query decomposition into existing retrievers on our newly-introduced \dataset{}, a large challenging dataset of real-world, tip-of-the-tongue queries for books.

\bibliography{anthology,custom}
\bibliographystyle{acl_natbib}

\appendix

\end{document}

%% file: tables/allexamples.tex
\begin{table*}[!ht]
  \centering
  \small
  \renewcommand{\arraystretch}{2} %
  \begin{tabular}{p{7cm}p{3cm}p{2cm}p{2cm}}
    \toprule
    \textbf{Query-Document} & \textbf{Title} & \textbf{Date} & \textbf{Cover} \\
    \midrule
    \emph{Query:} I think I saw this in a used store once and I remember saying to my new husband \" my daughter use to read that book to her little brother\" and it's funny because on the outside cover is a little girl reading a book to her little brother. It's called.....my book, or my story, or something simple like that. It would be about 15 or more years old. The girl was blond and the boy brunet....I think!!!! Inside was the cutest little sentences and my kids use to do what each page said. \"This is my nose\"... \"These are my eyes\".... Things like that. I'd love to see that book again....thank you!! & \emph{Clue:} It's called.....my book, or my story, or something simple like that. & \emph{Clue: 15 years old (2006 or earlier)}  & \emph{Clue:} The outside cover is a little girl reading a book to her little brother. \\
    \emph{Description: Glossy pictorial hardcover no dust jacket. 2001 7.75x9.13x25. GUIDE FOR PARENTS WITH PICTURES, HOW TO TEACHING CHILDREN READING.} & \emph{Field:} My First Book & \emph{Field:} First published September 1, 1984  & \begin{tabular}{@{}l@{}}
          \emph{Field:} \\ \includegraphics[width=2cm]{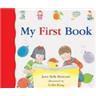}
    \end{tabular}  \\
    \hdashline
    \emph{Query:} BOOK, SPOILERS: i read the book just last year (2019) around september or october. i don’t think the book is older than maybe 2010. if i remember correctly the girls brother (donor of the heart) dies at a beach when he falls off a cliff during a race they had. also the boy who received the heart has a friend who’s a girl and she has cancer. he really likes drawing comic strips and he always drew her as a superhero with blue hair. i believe the cover had a pink heart on it and i think it was broken with a white background and the title of the book on or above the heart. & \emph{Clue:} The cover had a pink heart on it and i think it was broken with a white background and the title of the book on or above the heart. & \emph{Clue:} 2019 & \emph{Clue:} The cover had a pink heart on it and i think it was broken with a white background and the title of the book on or above the heart. \\
    \emph{Description:} Jonny knows better than anyone that life is full of cruel ironies. He's spent every day in a hospital hooked up to machines to keep his heart ticking. Then when a donor match is found for Jonny's heart, that turns out to be the cruellest irony of all. Because for Jonny's life to finally start, someone else's had to end.... & \emph{Field:} Instructions for a Second-hand Heart & \emph{Field:} First published December 1, 2017 & \begin{tabular}{@{}l@{}}
          \emph{Field:} \\ \includegraphics[width=2cm]{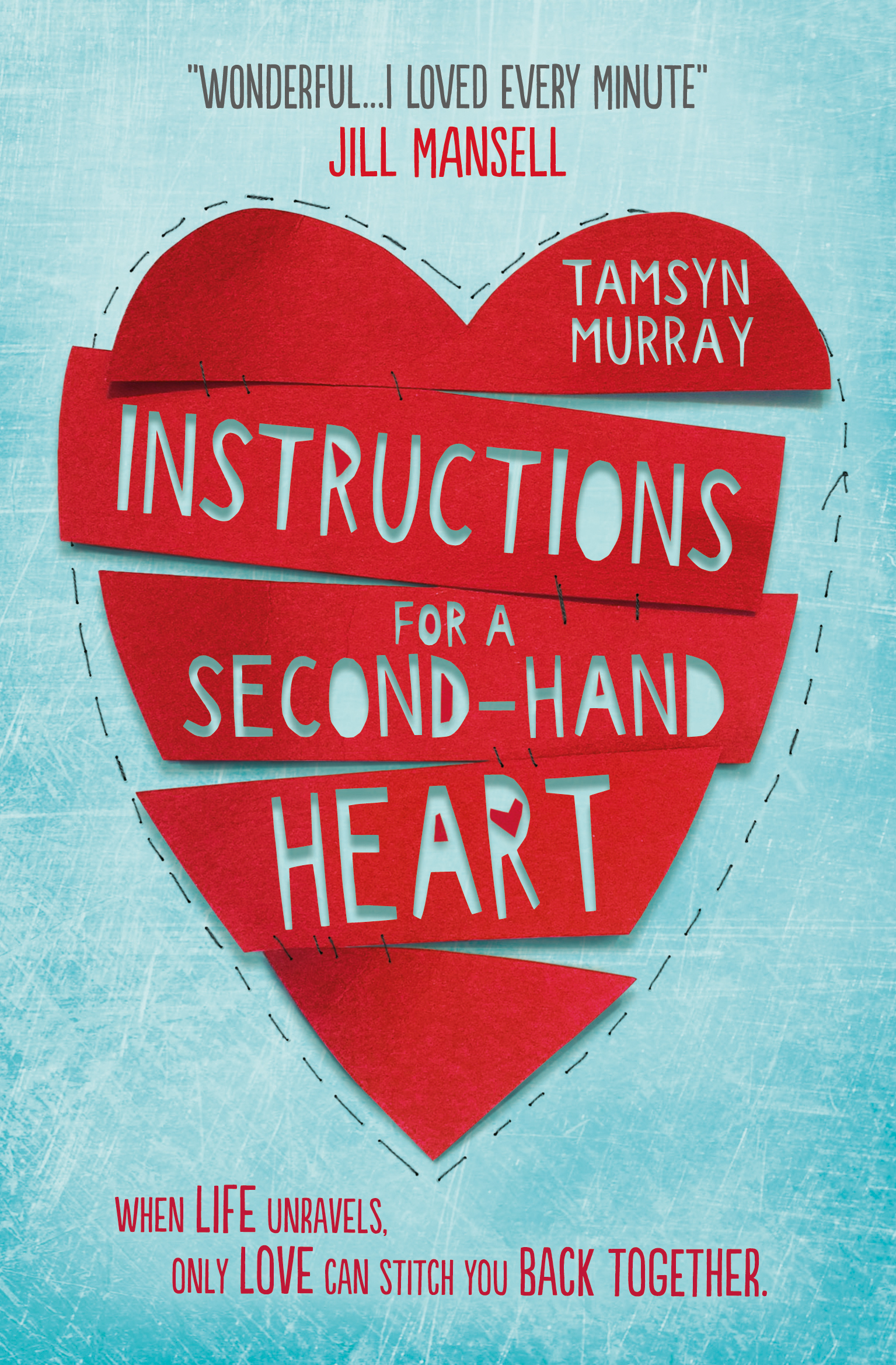}
    \end{tabular}   \\
    \hdashline
    \emph{Query:} The books is probably 11-20 years old. Written by a former journalist. Takes place in NYC. Involves a necklace by Marie Antionette. Something like SOCIAL GRACES, or SOCIETY GRACES. I read this probably in 2000? Thank you for your help. It's a great beach read. & \emph{Clue:} Something like SOCIAL GRACES, or SOCIETY GRACES. & \emph{Clue:} 2000 & \emph{Clue:} n/a \\
    \emph{Description:} When her husband of twenty years dies under mysterious circumstances, leaving his fortune--and Jo's position in society--to a mysterious French countess, Jo Slater, once one of New York's leading grande dames, comes up with an ingenious scheme to seek revenge designed to recoup her fortune and reclaim her "throne," with only a little murder standing in her way. Reprint. 75,000 first printing. & \emph{Field:} Social Crimes & \emph{Field:} First published June 12, 2002 & \begin{tabular}{@{}l@{}}
          \emph{Field:} \\ \includegraphics[width=2cm]{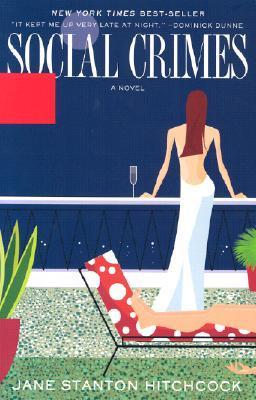}
    \end{tabular}   \\
    \bottomrule
  \end{tabular}
  \caption{Query-document pairs, their generated sub-queries or \emph{clues}, and corresponding gold document fields.}
  \label{tab:examples}
\end{table*}

%% file: tables/main-results.tex
\begin{table*}[h]
  \centering
  \small
    \begin{tabular}{lcccc} 
        \toprule 
        {\bf Model} 
        & {\bf Top 5} & {\bf Top 10} & {\bf Top 20} & {\bf Top 100} \\
        \midrule
        
        BM25~\citep{robertson-bm25-1997}    & 8.3 & 12.5 & 16.2& 22.5\\
        DPR~\citep{karpukhin2020dense} & 13.8  &  31.9  & 39.8 & 57.2\\

        CLIP~\citep{radford2021learning} & 1.9  & 2.8    & 3.5& 5.7\\
        Contriever~\citep{izacard2022unsupervised-contriever}  & 26.5  & 33.5 & 40.3 & 61.3 \\ \midrule
        
        Ours (Contriever+Date Expert only)  & 26.4  & 33.2 & 40.5 & 62.2 \\
        Ours  (Contriever+Image Expert only) & 27.2 & 34.9 & 42.0& 61.9\\
        Ours   (Contriever+Title Expert only) & 27.8 & 34.0 & 42.2& 61.2\\
        \textbf{Ours (Contriever+All Experts)} & \textbf{28.4} & \textbf{35.5} & \textbf{43.5} & \textbf{63.1} \\

        \bottomrule
    \end{tabular}
    \caption{Results on the test set of \datasetname. Metrics are Recall@$K$. The top half of models are single-retriever baselines; BM25, DPR, and Contriever all operate over book descriptions only, while CLIP operates over book covers only. All these baselines receive as input the full query. The bottom half of models make use of our query decomposition to obtain sub-queries and trained expert retrievers that operate over richer document representations.}
    \label{tab:main_results}
\end{table*}